\documentclass[a4paper]{llncs}
\usepackage[utf8]{inputenc}
\usepackage{cwdefs}
\usepackage{amssymb}
\usepackage{amsfonts}
\usepackage{amsmath}
\usepackage{xspace}
\usepackage{fancyvrb}
\usepackage{booktabs}
\usepackage[framemethod=TikZ]{mdframed}
\usepackage{xcolor}
\usepackage[export]{adjustbox}
\usepackage{tikz}
\usetikzlibrary{shapes.arrows}
\usepackage{enumitem}
\usepackage{graphicx}
\usepackage[hidelinks,pdftitle=KBSET]{hyperref}

\newcommand{\KBSET}{\mbox{\name{KBSET}}\xspace}
\newcommand{\KBSETL}{\name{\mbox{KBSET}/Letters}\xspace}
\newcommand{\KBSETN}{\name{\mbox{KBSET}/NER}\xspace}
\newcommand{\TEI}{TEI\xspace}
\newcommand{\LATEX}{\LaTeX\xspace}
\newcommand{\GND}{\name{GND}\xspace}
\newcommand{\SWIPL}{\name{SWI-Prolog}\xspace}
\newcommand{\GNUEMACS}{\name{GNU Emacs}\xspace}

\newcommand{\sym}[1]{\texttt{#1}}

\title{\KBSET\ -- Knowledge-Based Support for Scholarly Editing and Text
  Processing\\ with Declarative \LATEX Markup\\ and a
  Core Written in \SWIPL}

\author{Jana Kittelmann\inst{1} \and
Christoph Wernhard\inst{2}}

\institute{Martin-Luther Universität Halle-Wittenberg, Germany
  \and Berlin, Germany}

\begin{document}

\maketitle

\begin{abstract}
  \KBSET is an environment that provides support for scholarly editing in two
  flavors: First, as a practical tool \KBSETL that accompanies the development
  of editions of correspondences (in particular from the 18th and 19th
  century), completely from source documents to PDF and HTML
  presentations. Second, as a prototypical tool \KBSETN for experimentally
  investigating novel forms of working on editions that are centered around
  automated named entity recognition.  \KBSET can process declarative
  application-specific markup that is expressed in \LATEX notation and
  incorporate large external fact bases that are typically provided in
  RDF. \KBSET includes specially developed \LATEX styles and a core system
  that is written in \SWIPL, which is used there in many roles, utilizing that
  it realizes the potential of Prolog as a unifying language.
\end{abstract}

\section{Introduction}
\label{sec-introduction}

In the age of Digital Humanities, scholarly editing
\cite{plachta,sahle} involves the combination of natural language text
with machine processable semantic knowledge, typically expressed as
markup.  The best developed machine support for scholarly editing is
the XML-based TEI format \cite{tei:360}, a comprehensive markup
language for all sorts of text, mainly targeted at rendering for
different media and extraction of metadata, which is achieved through
semantics-oriented or declarative markup.  Recent efforts stretch \TEI
by aspects that are orthogonal to its original \name{ordered hierarchy
  of content objects (OHCO)} text model, through support for entities
like \name{names, dates, people, and places} as well as structuring
with \name{linking, segmentation, and alignment} \cite[Chap.~13
  and~16]{tei:360}.  Also ways to combine \TEI with Semantic Web
techniques, data modeling and ontologies are investigated
\cite{eide:2015}.  In accord with these directions we observe a number
of apparently open desiderata for the support of scholarly editing in
today's practice and in future perspective, which we explicitly
address with our environment \KBSET (\textbf{K}nowledge-\textbf{B}ased
\textbf{S}upport for Scholarly \textbf{E}diting and \textbf{T}ext
Processing):
\begin{enumerate}
\setlength{\parskip}{0.5ex}
  
\item \label{req-markup} It should be possible for users from the application
  domain to \emph{create, review, validate and maintain source documents} of
  the edition project. That is, documents with annotated text, with metadata,
  and with data on relevant entities such as persons and locations. Text
  markup should be exposed to the users as far as it is relevant and
  interesting for the application field. Source documents must be stored and
  versioned.  Since source texts with XML markup are hardly readable, in the
  TEI/XML approach typically an additional user-interface layer is added to
  the workflow, where apparently only a single -- non-free -- software system
  is suitable.\footnote{The \name{Oxygen XML Editor}. See also
    \url{https://en.wikipedia.org/wiki/Comparison_of_XML_editors}, accessed
    Nov 19 2019.} On the other hand, outside the Humanities, with \LATEX the
  direct use of text with markup is widespread, well supported by many free
  tools and supplemented by numerous free packages of high
  quality.\footnote{In fact, \cite[Sect.~iv]{tei:360} notes that \textit{``the
      TEI encoding scheme itself does not depend on this language}
    [XML]\textit{; it was originally formulated in terms of SGML (the ISO
      Standard Generalized Markup Language), a predecessor of XML, and may in
      future years be re-expressed in other ways as the field of markup
      develops and matures.''}}
  
\item \label{req-reproducible} It should be possible to generate
  \emph{high-quality} print and hypertext presentations in a
  \emph{reproducible} way, based on published source documents created in the
  edition project as well as additional documents and programs that are freely
  available and can be precisely identified.

\item \label{req-intermediate} Not just ``final'' presentations should be
  well-supported but also \emph{internal tools} for developing the scholarly
  edition and \emph{intermediate presentations} used there should be of high
  quality.  This is in particular relevant as many edition projects take
  several years.
  
\item \label{req-external} It should be possible to couple object text with
  associated information in ways that are \emph{more flexible than in-place
    markup}: It may be convenient to maintain text annotations separately from
  the commented text sources. Markup can be by different authors,
  automatically generated, or for some specific purpose. Some queries and
  transformations should remain applicable also after changes of the markup.

\item \label{req-fuzzy} It should be possible to incorporate advanced
  semantics related techniques that inherently deliver result that are
  \emph{fuzzy}, \emph{imprecise}, or \emph{incomplete}. For example, named
  entity recognition or tools for statistics-based text analysis.

\item \label{req-link} \emph{Linking with external knowledge bases} should be
  supported.  These include results of other edition projects as well as large
  fact bases such as authority files like \name{Gemeinsame Normdatei
    (GND)},\footnote{\label{footnote-gnd}\url{http://www.dnb.de/gnd}. The \GND
    is maintained by the German-speaking library community and contains
    information about various entities, in particular about more than 11
    million persons in more that 160 million fact triples.  It is in the
    public domain (CC0) and can be downloaded as an RDF/XML document whose
    decompressed size is more than 18~GB.}  metadata repositories like
  \name{Kalliope},\footnote{\url{http://kalliope-verbund.info}.}  domain
  specific bases like \name{GeoNames}, or aggregated bases like \name{YAGO}
  \cite{yago2} and \name{DBpedia} \cite{dbpedia}.
  
\item \label{req-logic} A digital edition project involves, more or less
  explicitly, the creation of data, in other words, the assertion of facts
  about relevant entities like persons, locations, dates, events and units of
  text such as, for example, letters as components of a correspondence, or
  distinguished positions in texts. Such data can be project-specific or
  obtained through combination with external fact bases. As a result of an
  edition project, such \emph{data should be made explicit and accessible} in
  a way that facilitates to associate with them \emph{machine processable
    semantics}, that is, meanings based on some logic that is supported by
  tools from automated reasoning and knowledge processing.  Ontology reasoning
  in description logics is important here, but, by itself, not sufficient, as
  classification seems not a main operation of interest in the field. The \GND
  fact base on persons, institutions and works, for example, gets by with a
  quite small ontology of 64 classes.
\end{enumerate}
\KBSET approaches these desiderata successfully through the involvement of two
technologies: \LATEX and Prolog. More specifically, we defined a dedicated
small set of descriptive markup elements that is tailored to the application
domain, in our case the scholarly edition of correspondences of the 18th and
19th century, in the form of \LATEX commands and environments, and use \SWIPL
\cite{swiprolog} as a \emph{single environment and language} to implement all
tasks that involve parsing and composition of documents and fact bases in
various formats, querying with respect to documents and fact bases, and
evaluation of complex application constraints.

The current version of \KBSET supports two flavors of application: The first,
\KBSETL, is a practical environment for scholarly editions of
correspondences. Implemented support covers in particular editions of
correspondences from the 18th and 19th century in German language. The second,
\KBSETN, is a prototype system that allows to experiment with various advanced
features centered around named entity recognition.  \KBSETL is currently
applied in a large project, the edition of the correspondence of philosopher
and polymath Johann Georg Sulzer (1720--1779) with author, critic and poet
Johann Jakob Bodmer (1698--1783), which will be published in print as
\cite[Vol.~10]{sulzer:gs} in summer 2020. Including annotations and indexes,
it spans about 2000 printed pages.  The online HTML edition, also generated
with \KBSETL from the same sources, will be published in parallel.  In
addition, \KBSET is applied in a long-term project,
\href{http://www.sulzer-digital.de}{\name{www.sulzer-digital.de}}, a digital
representation of Sulzer's complete correspondence, edited successively with
\KBSET.
To illustrate the use of \KBSETL, the distribution of \KBSET includes the
edition of a small correspondence.  For \KBSETN it includes as an example a
draft edition of a 19th century book.  \KBSET is available as free software
from its home page
\begin{center}
\url{http://cs.christophwernhard.com/kbset.}
\end{center}
The 2016 version of \KBSETN was presented at DHd~2016
\cite{Kittelmann:Wernhard:DHd:2016}.  The Sulzer-Bodmer edition project and
its use of \KBSETL, as well as related further interdisciplinary research
topics, are described (in German) in
\cite{Kittelmann:Wernhard:halle:2019}. Some components of \KBSETL were derived
from an earlier collaboration of the authors,
\href{http://www.pueckler-digital.de}{\name{www.pueckler-digital.de}}
\cite{puecklerdigital}.

The rest of the paper is structured as follows: In
Sect.~\ref{sec-kbset-letters} we describe \KBSET/\linebreak\name{Letters}, the
environment for preparing scholarly editions of correspondences in practice,
and in Sect.~\ref{sec-kbset-ner} the more experimentally oriented \KBSETN
flavor of \KBSET centered around named entity recognition.  We conclude the
inter-disciplinary paper in Sect.~\ref{sec-discussion} with discussions of the
\KBSET environment from three different perspectives: Tools for scholarly
editing, the role of \SWIPL as a unifying practical technology, and some
encountered issues that might be of interest for future research on
logic-based knowledge processing.

\section{\KBSETL}
\label{sec-kbset-letters}

The \KBSETL environment is at its current state of development adequate for
scholarly editions of correspondences from the 18th and 19th century that are
in German language and where the edited texts are represented in a
character-preserving (\name{zei\-chen\-ge\-treu}) but not position-preserving
(\name{posi\-tions\-ge\-treu}) way.

\subsection{Descriptive Application-Specific Markup in \LATEX Notation}


\definecolor{boxbg}{rgb}{0.92,0.92,0.92}
\definecolor{arrowbg}{rgb}{0.8,0.8,0.8}

\mdfdefinestyle{figbox}{%
     outerlinewidth=0.1pt,
    roundcorner=5pt,
    innertopmargin=7.8pt,
    innerbottommargin=7.8pt,
    innerrightmargin=7.8pt,
    innerleftmargin=7.8pt,
    backgroundcolor=boxbg}

\newenvironment{figbox}[2]
               {\begin{minipage}[t]{5.90cm}%
                   \begin{mdframed}[style=figbox]%
                     \small\raggedright%
                       {\centering \rule{0pt}{1.7ex}\textbf{#1: #2}\\[3pt]}\par}
               {\rule[-0.58ex]{0pt}{2.0ex}\end{mdframed}\end{minipage}}

\newcommand{\nlf}{\\[4pt]}
\newcommand{\nlfl}{\\[2pt]}
\newcommand{\xbull}{\raisebox{0.1ex}{\textbullet\hspace{0.5em}}}
\newcommand{\xbullspc}{\hphantom{\xbull}}

\newcommand{\figarrow}{
    \begin{tikzpicture}
      \node[scale=1.0,single arrow,draw=none,fill=arrowbg,minimum height=0.7cm,shape border rotate=270] at (0,0) {};
    \end{tikzpicture}
}

\begin{figure}
\noindent
\textbf{\large Inputs}
\vspace{-20pt}

\noindent
\hfill
\begin{figbox}{I1}{Object Text Documents}
  \textit{Format:} \LaTeX{} with domain-specific
  descriptive markup\nlf
  \textit{Tool:} \GNUEMACS
\end{figbox}
\hspace*{\fill}

\vspace{0.25cm}

\noindent
\begin{minipage}[t]{0.5\textwidth}

\hspace*{\fill}
\begin{figbox}{I2}{Annotation Documents}
  Annotations that are maintained externally from
  object text\nlf
  \textit{Format}, \textit{Tool}: Same as for object text documents
\end{figbox}
\hspace*{0.00cm}

\vspace{0.25cm}

\hspace*{\fill}
\begin{figbox}{I4}{Assistance Documents}
  To configure and adjust \KBSET\nlf
  \textit{Format}: \KBSET-specific, Prolog-readable\nlf
  \textit{Tool}: \GNUEMACS
\end{figbox}
\hspace*{0.00cm}
\end{minipage}
\begin{minipage}[t]{0.5\textwidth}

\hspace*{0.00cm}  
\begin{figbox}{I3}{Application Fact Bases}
  About, e.g., persons, works, locations; bibliography\nlf
  \textit{Formats:} Prolog, \LaTeX{} markup, BibLaTeX\nlf
  \textit{Tools}: \GNUEMACS, \name{JabRef}
\end{figbox}

\vspace{0.25cm}

\hspace*{0.00cm}
\begin{figbox}{I5}{Large Imported Fact Bases}
  E.g., \GND, \name{GeoNames}, \name{Yago}, \name{DBPedia}\nlf
  \textit{Formats:} E.g., RDF/XML, CSV
\end{figbox}  
\end{minipage}

\vspace{5pt}

\hfill\figarrow\hspace*{\fill}

\vspace{-15pt}

\noindent
\textbf{\large Core system}

\noindent
\begin{minipage}[t]{0.5\textwidth}

\hspace*{\fill}
\begin{figbox}{C1}{Text Combination}
  \xbull Reordering object text fragments,\\
\xbullspc e.g., letters by different writers in\\
\xbullspc chronological order\nlfl
\xbull Merging with external annotations\nlfl
\xbull Merging with automatically\\
\xbullspc generated annotations
\end{figbox}
\hspace*{0.00pt}

\vspace{0.25cm}

\hspace*{\fill}
\begin{figbox}{C3}{Named Entity Identification}
  Persons, locations, dates
\end{figbox}
\hspace*{0.00pt}
\end{minipage}
\begin{minipage}[t]{0.5\textwidth}

\hspace*{0.00pt}  
\begin{figbox}{C2}{Consistency Checking}
  E.g.,\ for void
  entity identifiers, insufficient or implausible date specifications, duplicate
  entries in fact bases
\end{figbox}

\vspace{0.25cm}

\hspace*{0.00pt}
\begin{figbox}{C4}{Register Generation}
\xbull Indexes for print presentations\nlfl
\xbull Overview and navigation\\\xbullspc documents for Web presentation
\end{figbox}
\end{minipage}

\vspace{5pt}

\hfill\figarrow\hspace*{\fill}

\vspace{-15pt}

\noindent
\textbf{\large Outputs}

\noindent
\begin{minipage}[t]{0.5\textwidth}

\hspace*{\fill}
\begin{figbox}{O1}{$\!$Display of Identified Entities}
  \textit{Tool:} \GNUEMACS
\end{figbox}
\hspace*{0.00pt}
\end{minipage}
\begin{minipage}[t]{0.5\textwidth}
\hspace*{0.00pt}  
\begin{figbox}{O2}{Fact Bases}
    \textit{Formats:} E.g., RDF/XML, Prolog
\end{figbox}
\end{minipage}
  
\vspace{0.25cm}

\noindent
\begin{minipage}[t]{0.5\textwidth}

\hspace*{\fill}
\begin{figbox}{O3}{Print-Oriented Presentation}
  \textit{Formats:} \LaTeX, PDF
\end{figbox}
\hspace*{0.00pt}
\end{minipage}
\begin{minipage}[t]{0.5\textwidth}
\hspace*{0.00pt}  
\begin{figbox}{O4}{Web-Oriented Presentation}
    \textit{Format:} HTML
\end{figbox}
\end{minipage}

\caption{\KBSET: Overview on inputs, core system functionalities and outputs}

\label{fig-overview}

\end{figure}

Figure~\ref{fig-overview} shows an overview on \KBSET: Inputs, functionalities
of the core system that is implemented in \SWIPL, and outputs.  For creating a
scholarly edition of a correspondence, the inputs are documents with domain
specific markup expressed as \LATEX commands and environments, representing
object texts of the edition project, that is, letters, and annotations by the
editors that refer to the object texts, respectively (box~I1 and~I2 in
Fig.~\ref{fig-overview}).  The parsimonious set of declarative markup elements
\name{KBSET/Letters Markup}\footnote{A specification draft is available from
  the KBSET home page.}  is tailored to the requirements of such scholarly
editions. Through the specialization, creating the markup is perceived by
users as expressing statements of interest rather than a technical burden.
Through the \LATEX notation, the marked-up text remains fairly readable and
can be directly created by users with any text editor that supports \LATEX,
such as, for example, \GNUEMACS, which is free software and shown as
representative tool in the figure.
 
Letters and annotations are represented by \LATEX environments.  Here is an
example of a letter environment:

\begin{Verbatim}[fontsize=\small]
\begin{letter}{bs:1745-02-14}{bodmer}{sulzer}{zuerich}{14. Februar 1745}
... 
Der Hr.~\xperson{lange}{Pastor Lange von Laublingen}, hat mir, noch
 \xl{brief:lange}{ehe er den Brief von E~Hochedl. empfangen}, berichtet,
...
\end{letter}
\end{Verbatim}

\noindent
Identifier \sym{bs:1745-02-14} is declared to denote the represented letter.
Arguments of the \verb+\begin{letter}+ statement provide essential meta data:
Identifiers of writer, addressee and location, as well as the date in a human
readable but parsable form.  The tilde for non-breaking space is transferred
from \LATEX to the \KBSETL markup.  The phrase \name{Pastor Lange von
  Laublingen} is marked-up as denoting the person with identifier
\sym{lange}. Identifiers used here can be mnemonic as they are local to the
project.  The identifier \sym{brief:lange} is declared to denote the marked-up
occurrence of the phrase \name{ehe er den Brief von E~Hochedl. empfangen} in
the letter.  Its scope is the letter environment. The following example shows
an annotation environment:

\begin{Verbatim}[fontsize=\small]
\begin{annotation}{bs:1745-02-14}
  ...
  \ksection{Stellenkommentar}
  \begin{klist}
  \kitem{brief:lange} Der Brief Bodmers an Samuel Gotthold Lange ...
  ...
  \end{klist}
\end{annotation}
\end{Verbatim}

\noindent The annotation block is about the example letter above, associated
through the argument \verb+bs:1745-02-14+ of the \verb+\begin{annotation}+
statement.  In the annotation environment the identifiers like
\sym{brief:lange} that were locally declared in the letter environment are
re-activated for referencing. This permits a convenient way to express
annotations that refer to specific places in the text of letters
(\name{Stellenkommentare}).

Also fact bases can be written with special markup commands in \LATEX
notation. For example, the referenced person \sym{lange} can be declared with
the following statement:

\begin{Verbatim}[fontsize=\small]
\defperson{lange}{Lange, Samuel Gotthold (1711--1781)}
\end{Verbatim}

\noindent
Person names in these declarations must be compatible with the regularities
used by the \GND.\footnote{We do not demand in full the principles of the \GND
  for choosing \name{preferred} names, as \name{``Colombo, Cristoforo''} or
  \name{``Homerus''} is unusual in German texts.} They can be directly used in
indexes and, with years of birth and death, allow to automatically determine
the global \GND identifiers of persons represented in the \GND.  These global
identifiers make metadata maintained, for example, in the \GND and
\name{Wikipedia} available, relieving the edition project from the need to
replicate them explicitly.

So far, the user perceives the project as a collection of documents with
letters, annotations and fact bases in the specialized descriptive \LATEX
markup. Indeed, \KBSET provides an implementation of the specialized markup in
form of a \LATEX package that is sufficient to generate a PDF representation
of the letters and annotations with fairly high quality just by a pure \LATEX
workflow. In the result, letters and associated annotations are connected
through PDF hyperlinks.  References like \verb+\xperson{lange}{...}+ to
identifiers declared in a fact base are converted to index entries processed
by \name{xindy}. The bibliography is handled by \name{BibLaTeX}.  The involved
\LATEX processors already ensure validity and consistency of the documents to
some degree.

\subsection{From \LATEX to Prolog for
  Further Consistency Checking and Text Combination}

The \KBSET core system includes a \LATEX parser written in Prolog that yields
a list of items, terms whose argument is a sequence of characters represented
as atom, and whose functor indicates a type such as \name{word},
\name{punctuation}, \name{comment}, \name{command}, or \name{begin} and
\name{end of an environment}. A special type \name{opaque} is used to
represent text fragments that are not further parsed, such as \LATEX
preambles.  \LATEX commands and environments can be made known to the parser
to effect proper handling of their arguments.  The parser aims to be
practically useful, without claiming completeness for \LATEX in full.  It does
not permit, for example, a single-letter command argument without enclosing
braces. The parser is supplemented by conversions of parsing results to \LATEX
and to plain text.

So far, additional syntactic checks at parsing and various semantics-oriented
checks that are applied after the parsed documents are converted to Prolog
fact bases are implemented (box~C2 in Fig.~\ref{fig-overview}). Further ways
of consistency checking can be realized with respect to the generated HTML
documents discussed below in Sect.~\ref{sec-html}.

\begin{figure}
  \centering
  \includegraphics[width=0.483\textwidth,cfbox=lightgray 1pt 1pt]
                  {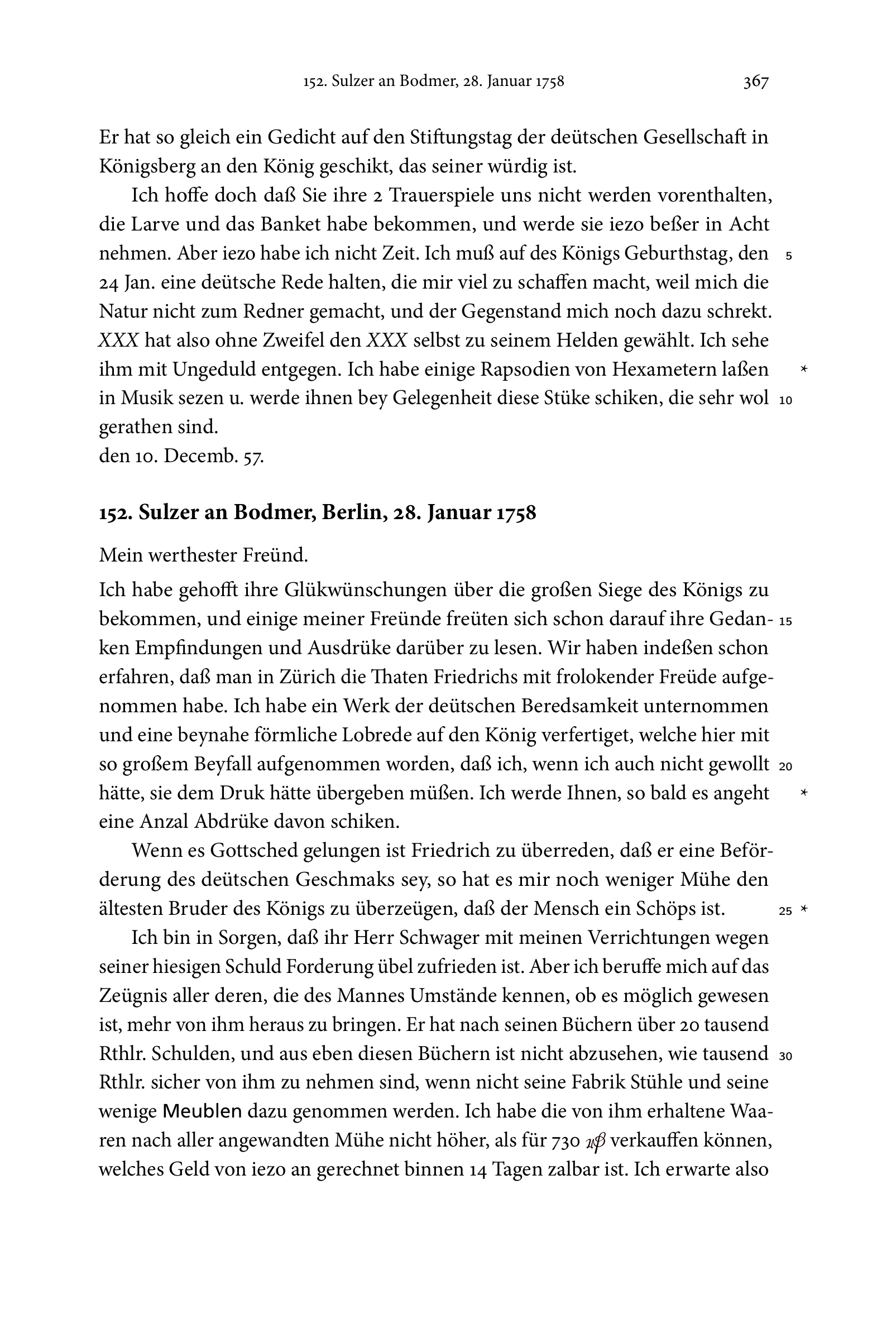}
  \hfill
  \includegraphics[width=0.483\textwidth,cfbox=lightgray 1pt 1pt]
                  {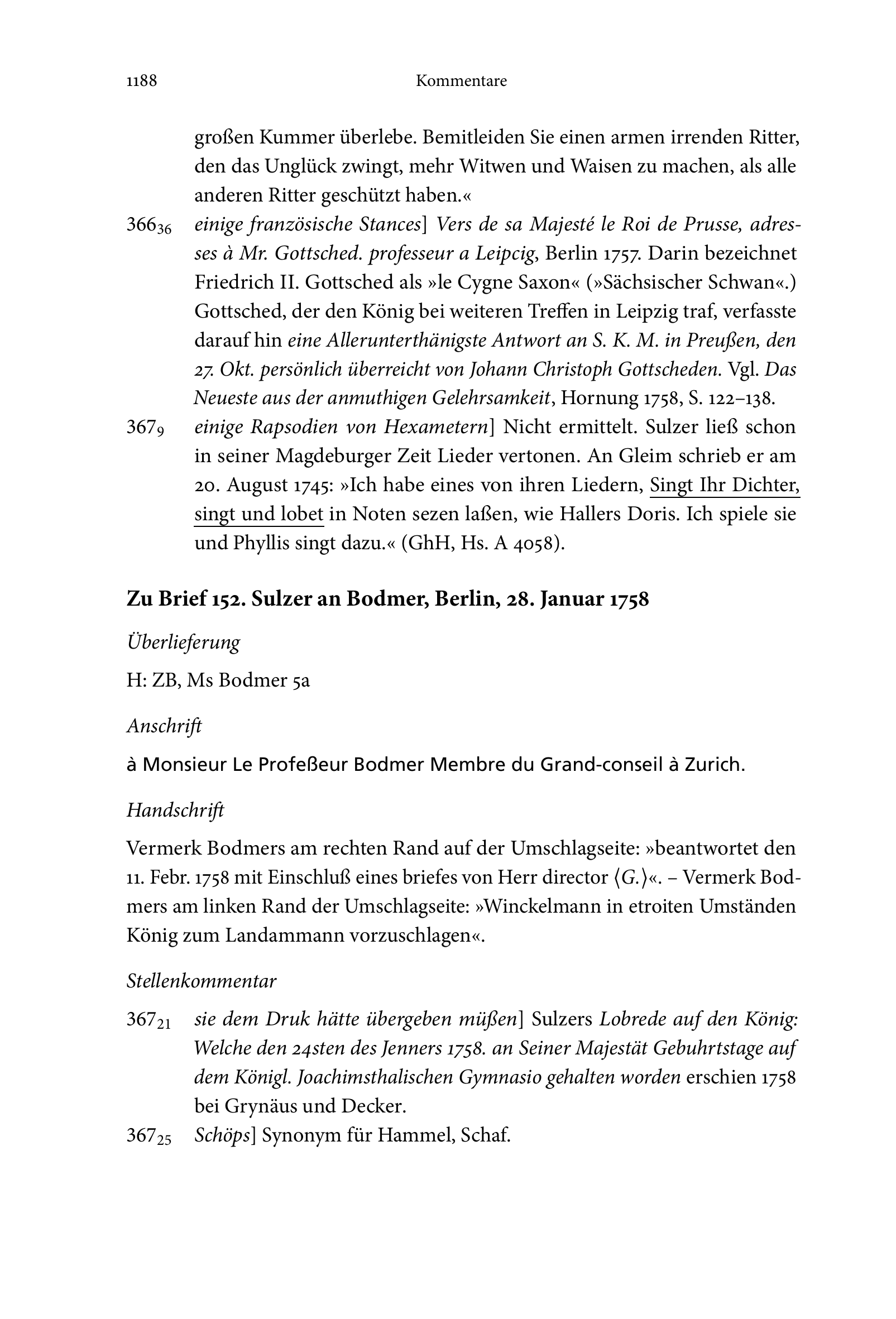}
  \caption{\KBSETL: PDF presentation, a letter and an annotation page.}
  \label{fig-pdf}
\end{figure}

Source documents with letters and with annotations are maintained in a large
edition project not necessarily in the same ordering and fragmentation in
which these should appear in presentations. Based on the parsed \LATEX, the
\KBSET core system can perform such rearrangements (box~C1 in
Fig.~\ref{fig-overview}) and write out generated \LATEX documents.  The
conventional \LATEX workflow applied to these generated documents then results
in high-quality PDF documents, which, depending on the configuration, are
suitable for publication in print or on-screen reading.\footnote{Before
  printing in high quality, \LATEX documents in general need manual
  adjustments in places that can not be handled satisfactorily by the automated
  layout processor.} Figure~\ref{fig-pdf} shows example output pages.

The functionalities for consistency checking and text combination are
available as Prolog predicates in a user interface module, and, for users that
do not want to interact with Prolog directly, with \name{Bash} shell scripts
that invoke \SWIPL.\footnote{In Microsoft Windows, these scripts can
  be called from the \name{Cygwin} shell.}

\subsection{HTML Presentation}
\label{sec-html}

The parsed source documents are converted to representations as Prolog
predicates, which form the basis for generating an HTML representation of the
scholarly edition.  In general, our Web presentation is designed to open-up
the edition, to make it easy to get an overview on the material and on the
supported navigation possibilities.

On the basis of the identifiers in the \LATEX-syntax source documents, URIs
for documents like letters and entities like persons and locations are
generated.\footnote{This requires a syntactic conversion as ``\texttt{:}'' has
  a special meaning in URIs.} These can be used as URLs of the respective
generated pages, which then can persistently represent the respective document
or entity with respect to the edition.  An HTML presentation of the project
bibliography is generated from the \name{BibLaTeX} sources via an invocation
of the \name{Biber} processor with options such that it produces an XML
representation of the processed bibliography that is then read into \SWIPL.
  
The Web presentation just uses static pages, in HTML5, with CSS3 and -- very
little -- JavaScript.  This makes the loading of pages fast, requires no
maintenance efforts, and facilitates the interaction with search engines,
general Web search engines as well as dedicated engines for the online
publication.

Some simple but useful means for navigation were realized: Letter pages have
links to a chronologically next and previous letter, with respect to the
writer and also with respect to the correspondence with the addressee. These
four links are always displayed at the same position in the page and thus
allow to quickly move within the letters by an author or in a correspondence.

\begin{figure}
  \centering
  \includegraphics[width=1.0\textwidth]
                  {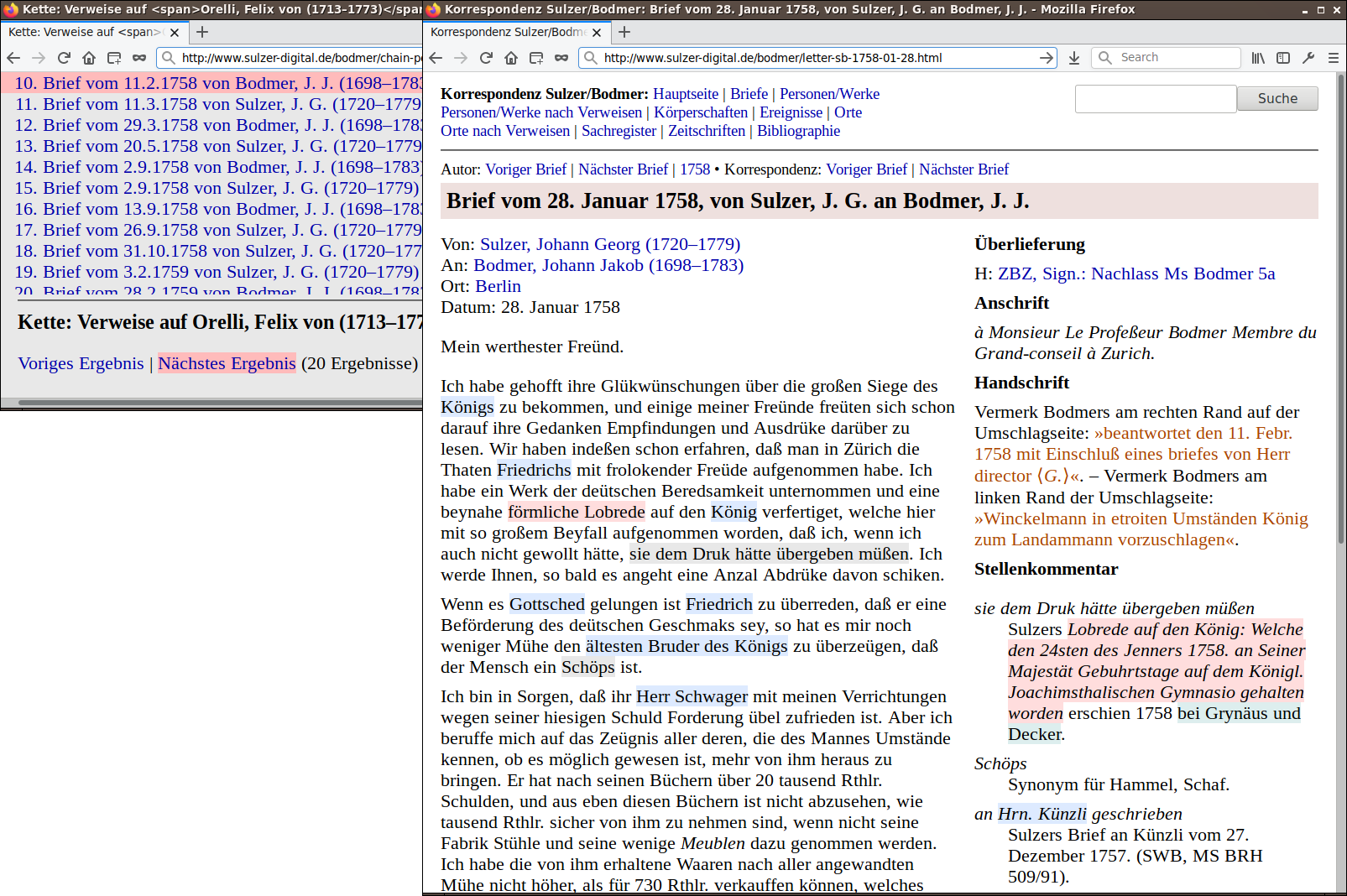}
  \caption{\KBSETL: HTML presentation, a letter and a chain page.}
  \label{fig-html}
  \vspace{-0.6cm} 
\end{figure}

Another realized useful navigation means is what we call \name{chains}
(\name{Ketten}), or, more explicitly, \name{result value chains}: The value of
a query is often a ``chain'', that is, an ordered set of entities, represented
as a series of links.  Navigating through such a chain is facilitated by a
special type of Web pages, \name{chain pages}, which just display the chain of
links but are invoked parameterized by an index into the chain. They scroll
their content automatically such that the indexed link appears at the top.  By
clicking at some link or a \name{next} button (for the indexed link) in the
chain window the respective linked document is opened in a different window,
and the index of the chain window is incremented (a \name{previous} button has
analogous effect). We actually use chains for a finite number of precomputed
queries of general interest such as the set of all letters in which a given
person is referenced, and whose results are also displayed on the respective
entity pages -- but are there less convenient to browse through.  Chain pages
are by default shown in a small pop-up window positioned top left on the
screen.  If possible, an existing chain window and an existing window for
displaying a page linked from a chain window are re-used. Our implementation
utilizes the CSS3 \texttt{target} attribute.  Figure~\ref{fig-html} shows an
example of a generated Web page representing a letter, accompanied by a chain
page.

\subsection{Access from Prolog and Export of Fact Bases}
\label{sec-access-from-prolog}

The advanced consistency checking and text combination, as well as the HTML
generation can be invoked via \name{Bash} shell scripts or directly from
Prolog.  The representation of the parsed source documents as Prolog
predicates underlying the HTML conversion can in principle also be applied for
other applications, such as conversion to further formats like RDF/XML and
TEI/XML, or to export fact bases as indicated with box~O2 in
Fig.~\ref{fig-overview}.  The plan is to specify a suitable set of Prolog
predicates such that editions can offer exported data, and also the parsed
text, for download.  Currently ways to produce RDF and XML on the basis of the
internal Prolog predicates are indicated with small examples in the source
code.

\section{\KBSETN}
\label{sec-kbset-ner}

While \KBSETL addresses the practical aspects of comprehensive scholarly
correspondence editions, the focus of \KBSETN is to explore experimentally
potential future directions of scholarly editing.  Specifically the
integration of techniques that return non-symbolic, fuzzy or incomplete
results, the utilization of large external fact bases from the library
community such as the \GND and from Semantic Web activities such as
\name{YAGO}, \name{DBPedia} and \name{GeoNames}, and ways to handle the
association of annotations with places in the object text that are not
explicitly marked as reference target. The functionality of \KBSETN can be
accessed from the Prolog interpreter or with menus and keyboard shortcuts from
\GNUEMACS.  A draft edition of \name{Geschichte der Reaction}, vol.~1, 1852,
by philosopher Max Stirner that has been created with these novel techniques
is included with the \KBSET distribution.

\subsection{Caching External Knowledge Bases for Access Patterns}
\label{sec-caches}

The inputs of \KBSETN include, aside of object texts and annotation documents
(boxes~I1 and~I2 in Fig.~\ref{fig-overview}), also large imported fact bases
(box~I5).  Before use, the configured fact bases, which are typically
available in Semantic Web formats like RDF/XML or as CSV tables, have to be
downloaded, parsed and preprocessed.  This can be done with a utility
predicate, but, as it may take several hours, for the example application also
a TAR archive with the results of the preprocessing can be downloaded from the
\KBSET home page.\footnote{Also the original fact bases used for the example
  application are archived on the \KBSET home page, as none of them has a
  persistent URL.}  The preprocessed fact bases are then loaded into the
Prolog system. At the first loading they are compiled into \SWIPL's
\name{quick-load} format. In that format our fact base with 12 million ternary
facts on persons born before 1850 extracted from the \GND takes 7 seconds to
load on a modern notebook computer.

\KBSET then accesses these data as Prolog predicates stored in main memory.
The indexing mechanisms of \SWIPL are utilized by maintaining predicates that
are adapted to the represented entities, such as persons or locations (in
contrast to generic triple predicates as might be suggested by the RDF
format), and to access patterns.  For example, a predicate for accessing data
about a set of persons via a given last name and another predicate for
accessing data about a person via a given \GND identifier.  We call these
predicates, which may be in part redundant from a semantic point of view,
\name{caches}. In the current implementation, the caches are in part computed
when preprocessing the fact bases and in part when loading them.  With this
approach, the system can evaluate the several 10.000s of queries against the
fact bases required for named entity recognition on the example document in a
few seconds.  Another useful feature is the semantics-based restriction of the
large fact bases at preprocessing them. Since our example edition is a book
from 1852, we keep of the \GND only the facts about persons born before 1850.

\subsection{Named Entity Identification}

Working with \KBSETN is centered around a subsystem for named entity
recognition, which detects dates by parsing as well as persons and locations
based on the \GND and \name{GeoNames} as gazetteers, using additional
knowledge from \name{YAGO} and \name{DBpedia}.  Persons can be detected in two
modes, characterized by names as well as by functional roles like \name{King
  of}, \name{Duke of} and \name{Bishop of}.  Differently from systems like the
\name{Stanford Named Entity Recognizer} \cite{ner:stanford}, \KBSETN does not
just associate entity types such as \name{person} or \name{location} with
phrases but attempts to actually \emph{identify} the entities, hence we also
speak of \name{named entity identification}.

The identification of persons and locations is based on single word
occurrences with access to a context representation that includes the text
before and after the respective occurrence.  Hence an association of
\emph{word occurrences} to entities is computed, which is adequate for indexes
of printed documents and for hypertext presentations, but not fully compatible
with \TEI, where the idea is to enclose a \emph{phrase that denotes an entity}
in markup.

\begin{figure}
  \centering
  \includegraphics[width=1.0\textwidth]
                  {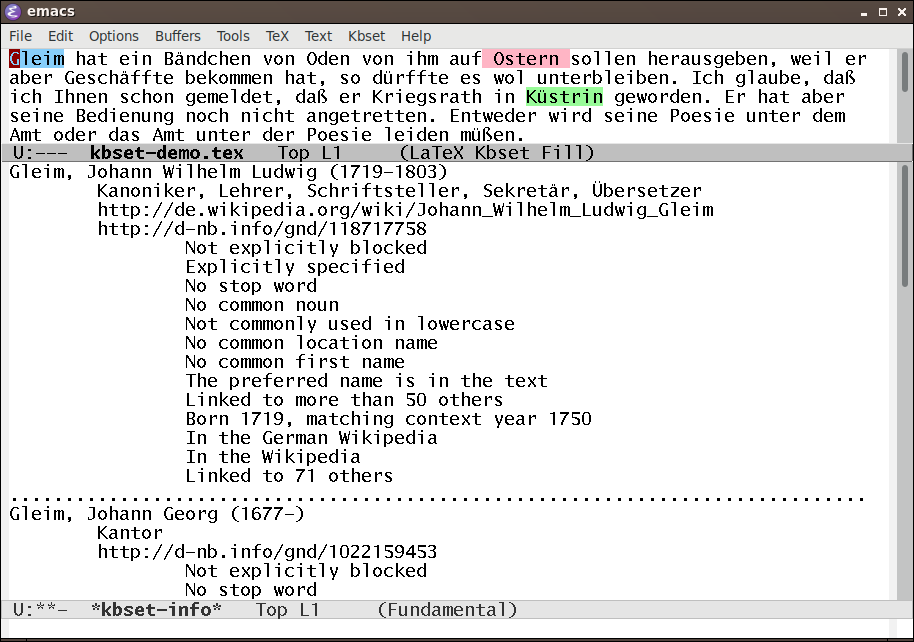}
  \caption{\KBSETN: Named entity identification from \GNUEMACS.}
  \label{fig-nei}
\end{figure}

Figure~\ref{fig-nei} shows the presentation of named entity identification
results in \GNUEMACS.  In the upper buffer, which contains the object text,
the system highlights words or phrases about which it assumes that they denote
a person, location or date. In the lower buffer additional information on the
selected occurrence of \name{Gleim} is displayed: Links to \name{Wikipedia}
and \GND, an explanation \emph{why} the system believes the entity to be a
plausible candidate for being referenced by the word occurrence, and an
ordered listing of lower-ranked alternate candidate entities.  Menus and
keyboard shortcuts allow to jump quickly between the highlighted text
positions with associated entities.

Aside of the presentation in \GNUEMACS, the results of named entity
identification can be output in different formats, in particular merged into a
\LATEX source document as annotations. In this merging process also external
annotation documents can be considered, where the positions to insert
particular annotations are abstractly specified, for example by some form of
text pattern. Further supported output formats of the named entity
identification results include the presentation as TEI/XML elements merged
into a source document, as a Prolog fact base, or, for identified locations,
as a CSV table that can be loaded into the \name{DARIAH-DE} geo browser.

The named entity identification is controlled by rules which can be specified
and configured and determine the evaluation of syntactic features matched
against the considered word, for example, \name{is-no-stopword} or
\name{is-no-common-sub\-stan\-tive}, and of semantic features matched against
candidate entities, for example, \name{is-in-wikipedia},
\name{is-linked-to-others-identified-in-context},
\name{has-an-occupation-mentioned-in-context}, or
\name{date-of-birth-matches-context}.  Evaluation of these features is done
with respect to the mentioned context representation, which includes general
information like the date of text creation and inferred information such as a
set of entities already identified near the evaluated text position.  Features
that are cheap to compute and have great effect on restricting the set of
candidate entities are evaluated first.  This allows, for example, to apply
named entity identification of persons on the 300-pages example book provided
with the system in about 7~seconds on a modern notebook computer.
Feature evaluation results are then mapped to Prolog terms whose standard
order represent their plausibility ranking. Information about the features
that contributed to selection of a candidate entity is preserved and used to
generate the displayed explanations shown in Fig.~\ref{fig-nei}.

\subsection{Assistance Documents}

The automated named entity identification produces incomplete, partially
incorrect and, by presenting a ranked list of plausible entities, fuzzy
results. Such results may be helpful for developing a scholarly edition but
should not remain in a released version. Hence, there must be a possibility to
adapt them. This can be done in \KBSETN with configuration files, so-called
\name{assistance documents} (box~I4 in Fig.~\ref{fig-overview}).  These
specify the complete configuration of \KBSETN, including the URLs of the
external fact bases, the preprocessing required to use them, and how to bias
or override automated inferencing in named entity identification.  The idea
for the latter is that the user, instead of annotating identified entities
manually, lets the system do it automatically and mainly gives hints in
\emph{exceptional} cases, where the automatic method would otherwise not
recognize an entity correctly.  That method was used in the example document
supplied with \KBSET.

In the assistance document the explicit appearance of technical identifiers
such as the identifiers from the \GND should be avoided.  This is achieved by
permitting to specify a person just by some attributes like name, year of
birth, and/or a profession. These specifiers are evaluated with the current
fact base of the system, that is, essentially the \GND.  In contexts where a
unique person must be designated an error is signaled if no or several persons
match the specifier. (Of course, this method is not stable against importing
an extended version of the \GND.) In addition, context properties can be
specified that characterize when the biasing should be applied.  Other options
are to register persons that can not be found in the \GND and to supplement
attributes of persons in the \GND. Also simple syntactic exclusions, for
example that a certain word should not denote a person or location can
be specified. Here is an excerpt from the assistance document for the included
example. It ``assists'' the automated named entity identifier in
distinguishing two persons named \name{Tacitus}, classifying \name{Starcke}
(the printer of the book) as a person and identifying him, as well as in
identifying the person referenced in the text as \name{Herzog von Luxemburg}:

\begin{Verbatim}[fontsize=\small]
entity(person,
       [name='Tacitus',
        professionOrOccupation='Historiker'],
       [near_word_in=['Römern']]),
entity(person,
       [name='Tacitus',
        variantNameForThePerson='Tacitus, Römisches Reich, Kaiser'],
       [near_word_in=['Adel']]),
entity(person,
       [name='Starcke, Johann Friedrich',
        professionOrOccupation='Drucker'],
       [near_word_in=['Druck']]),
supplement(person,
           [name='Joseph II., Heiliges Römisches Reich, Kaiser'],
           [biographicalOrHistoricalInformation
            =lang(de,'Herzog von Luxemburg (1765-1790)')]),
\end{Verbatim}

\noindent
Like Prolog program files, assistance documents can be re-loaded, which
effects updating of the specified settings. Thus, the named entity
identification of \KBSETL can be improved in an iteration of adjustments of
the assistance documents and reviewing the effects in the \GNUEMACS
presentation.

This mode of interaction has, however, in the Sulzer-Bodmer edition project
only been used in occasional cases.  Each letter has there been transcribed and
extensively commented by scientists, where the manual entity tagging emerged
as a by-product. The automated named entity identification has been applied in
special situations such as initializing the tagging of locations, examining
and completing the manual tagging of persons, and generating auxiliary fact
bases that map the project-local entity identifiers to global ones from the
\GND and \name{GeoNames}.

\section{Discussion}
\label{sec-discussion}

We conclude this inter-disciplinary paper with discussions of the \KBSET
environment from three different perspectives.

\subsection{\KBSET in the World of Tools for Digital Scholarly Editing}

\KBSET has been designed and written within the paradigm of programming and
creating mechanizable formalizations in Artificial Intelligence,\footnote{In
  the sense of the discipline \name{Artificial Intelligence}, not as
  synecdoche for its subfield \name{Machine Learning}.} which is considered
there as an integral component of the research activity.  The creation of text
with markup (\LATEX) is daily routine for researchers in computer science in
general, as well as in numerous further fields.

In contrast, in the Humanities the use of formally defined languages entered
in the last decade largely from the outside, with the requirement to make
publicly funded results openly available in the HTML-based Web.
Customizations of TEI/XML seemed the format of
choice.\footnote{\label{foot-dfg}See for example the DFG (German Research
  Foundation) document \name{Förderkriterien für wissenschaftliche Editionen
    in der Literaturwissenschaft, Ausgabe 11/2015},
  \url{https://www.dfg.de/download/pdf/foerderung/grundlagen_dfg_foerderung/informationen_fachwissenschaften/geisteswissenschaften/foerderkriterien_editionen_literaturwissenschaft.pdf}.}
Hence the creation of TEI/XML documents became a component of the scholarly
editing workflow.\footnote{Scholarly editions of correspondences that offer an
  openly available TEI/XML presentation include \name{Alfred-Escher
    Briefedition} (\url{https://www.briefedition.alfred-escher.ch}),
  \name{Briefe und Texte aus dem intellektuellen Berlin um 1800}
  (\url{https://www.berliner-intellektuelle.eu}), \name{Digitale Edition der
    Korrespondenz August Wilhelm Schlegels}
  (\url{https://august-wilhelm-schlegel.de}), \name{hallerNet}
  (\url{http://hallernet.org}), and \name{edition humboldt digital}
  (\url{https://edition-humboldt.de}).}  However, this should not be
misunderstood as equating the creation of digital editions to working with
TEI/XML.  The text represented in TEI/XML documents is hardly readable and the
computational treatment of TEI/XML usually requires familiarity with several
dedicated transformation and query languages, such that edition projects are
typically large undertakings that are accompanied by support from a
specialized IT department, which mediates between the formal languages and the
researcher.  Observe that this is quite different from research in Artificial
Intelligence, where the researcher herself creates mechanizable formalizations
and programs.  Where should the Digital Humanities go?

As TEI/XML is a general scheme for encoding all sorts of text, \name{``it is
  almost impossible to use the TEI schema without customizing it in some
  way''} \cite[Sect.~23.3]{tei:360}.  Applications such as scholarly editions
of letters typically use project- or organization-specific customizations.
Such customizations should be formalized in a schema language and explained in
an informal document, both of which should be made accessible with the digital
edition (\cite[Sect.~23.4]{tei:360} gives very specific notions of this, even
claimed to be presuppositions for calling a document \name{TEI-conformant}).
Unfortunately, in current practice such schema specifications and
documentations for digital editions are only rarely made easily
accessible.\footnote{Actually, the authors were (in November 2019) not able to
  find any correspondence edition where a formal specification of the used
  customized schema is referenced from the TEI/XML documents or specified on
  the Web site.  Informal edition guidelines can be found, for example, on the
  Web sites of \name{Alfred-Escher Briefedition}, \name{Briefe und Texte aus
    dem intellektuellen Berlin um 1800} and \name{hallerNet}.}  The suggested
way to associate a TEI/XML document instance with a schema by the
\verb+xml-model+ processing instruction \cite[Sect. v.7.2]{tei:360} seems not
used at all.\footnote{The well-intentioned postulation \name{``Um die
    Austauschbarkeit und Nachnutzung zu ermöglichen, werden die
    projektspezifisch verwendeten XML-Elemente und Attribut-Wert-Paare im
    TEI-Header dokumentiert''} in the DFG document mentioned in
  footnote~\ref{foot-dfg} can technically not refer to the \sym{teiHeader}
  element.}

Writing a conversion from \KBSETL source documents to some customization of
TEI/XML is an easy task based on the extraction process implemented for the
HTML transformation. The markup in \LATEX-syntax is there available in parsed
form, metadata appear as Prolog predicates, and routines for converting
identifiers are already implemented.  A module in \KBSET illustrates the
concrete proceeding for XML and RDF conversions of metadata. In fact, a
conversion to a TEI/XML customization is much simpler than the HTML
translation included in \KBSET.
It is not yet implemented for the reason that, so far, it seems difficult to
identify a particular formally defined TEI/XML customization for
correspondences for which interesting tools or services are openly available,
for example, to generate further presentations or for integration with other
editions.

In the light of the standardization efforts via TEI/XML, \KBSET can be taken
as a user-friendly and economic environment for developing scholarly editions
that approaches compliance with the desiderata described in the introduction.
The generation of a representation in some customized TEI/XML format for
interchange and archival is a marginal feature that is easy to add.  In the
long run, variations of the \KBSET markup language should perhaps be adapted
to reflect some suitable TEI/XML customizations more explicitly, or even be
considered as realizations of TEI customizations in \LATEX-syntax.

Vice versa, \KBSETL can also be taken as a tool for generating presentations.
It is not difficult to translate a representation of a correspondence in a
TEI/XML customization to the \KBSETL markup (this can be implemented on the
basis of the term representations of documents obtained from the XML parser of
\SWIPL) such that the PDF and HTML presentations offered by \KBSETL become
available.  Since, as already mentioned, projects use different and hardly
documented TEI/XML customizations it is expected that the translations need to
be project-specific and some trial-and-error is involved in the development.

\KBSET is free software. It depends only on a \TeX\ distribution (it has been
tested with \name{TeX Live}) and on \SWIPL, both of which are also free
software, platform independent, and, moreover, mature, stable and widely used
such that the current implementation of \KBSET can be expected to operate also
with future releases of these environments.\footnote{Some of the functions of
  \KBSET can be invoked in addition from \name{Bash} shell scripts. A
  \name{Bash} shell can be presupposed on Unix-like platforms and can be
  added, for example with \name{Cygwin}, to Microsoft Windows platforms.}

The sources of an edition project like the Sulzer-Bodmer correspondence can be
published and archived together with the used version of \KBSETL.  The
following functionalities are then freely available, through the stability and
platform independence of \LATEX and \SWIPL also in the foreseeable future:
Generation of various high-quality PDF and HTML representations, generation of
fact bases in Prolog representation,\footnote{Considering that there is an ISO
  standard for Prolog, such fact bases are actually in a \emph{standardized}
  format. However, the ISO standard for Prolog is only with respect to ASCII
  encoding. Modern implementations like \SWIPL support UTF-8.}  and the
representation in some TEI/XML customization (which still needs to be
implemented). Moreover, if users want to improve or extend these
functionalities, \KBSETL is available as a concrete and working free software
environment to begin with.

The use of \KBSET with other languages than German is supported to some
degree: All input documents created for \KBSET are encoded in UTF-8.  The
\GNUEMACS user interface of \KBSETN can be configured to English or
German. Some of the word lists included in the implementation are, however, so
far provided only for German.  Also the presentation templates of \KBSETL are
currently only in German. The \name{BibLaTeX} configuration included currently
with \KBSETL is based on practices of the Humanities in Germany, but it is no
problem to replace it with a different configuration.

\subsection{\SWIPL as a Unifying Practical Technology} 

The core system of \KBSETL is written in \SWIPL, which realizes the potential
of Prolog as a unifying language. As noted on the \SWIPL home
page,\footnote{\url{https://www.swi-prolog.org/features.html}, accessed Nov 21
  2019.} it considers Prolog \name{``primarily as glue between various
  components.  The main reason for this is that data is at the core of many
  modern applications while there is a large variety in which data is
  structured and stored. Classical query languages such as SQL, SPARQL, XPATH,
  etc. can each deal with one such format only, while Prolog can provide a
  concise and natural query language for each of these formats that can either
  be executed directly or be compiled into dedicated query language
  expressions. Prolog's relational paradigm fits well with tabular data
  (RDBMS), while optimized support for recursive code fits well with tree and
  graph shaped data (RDF).''} The particular roles of Prolog, and in particular
\SWIPL, for \KBSET can be compiled as follows:

\begin{enumerate}
  \setlength{\parskip}{0.5ex}
  
\item \name{Declarative representation mechanism for relational fact bases.}
  As outlined in Sect.~\ref{sec-access-from-prolog}, we convert the document
  sources created in scholarly edition projects and large external fact bases
  to an intermediate representation as Prolog predicates, which are then used,
  for example, to generate HTML pages, but are also available for other
  purposes, including export as fact bases or interactive querying on the
  Prolog shell.  The declarative view brings \emph{semantics} into the focus
  and offers a bridge to the wealth of semantics-based techniques for
  knowledge representation and knowledge-based reasoning, in particular
  deductive databases, model- and answer-set computation, first-order theorem
  proving, and ontology reasoning.
  
\item \name{Efficient representation mechanism for relational fact bases.}  We
  utilize the predicate indexing facilities of \SWIPL's with predicate caches
  that are specialized to access patterns as outlined in
  Sect.~\ref{sec-caches}.
  
\item \name{Query language.} The standard predicates \name{findall} and
  \name{setof} provide expressive means to specify queries in a declarative
  manner.  Complex tests and constructions can be smoothly incorporated, as
  query and programming language are identical, without much impedance
  mismatch.  Of course, queries written in Prolog can not rely on an
  optimizer, and have to be designed ``manually'' such that their evaluation
  is done efficiently.  A further important feature of Prolog is fast sorting
  based on a standard order of terms, which we quite often use to canonicalize
  representations of sets and is also the basis of our implementation of
  ranked answers in named entity identification.

\item \name{Representation mechanism for structured documents.}  As in Lisp,
  data structures are in Prolog by default terms that are print- and readable,
  a feature which is supplemented to ``non-AI'' languages often by XML
  serialization.  In our application context this is particularly useful as it
  allows to represent XML and HTML documents directly as Prolog data
  structures, that is, terms.

\item \name{Parser for XML and Semantic Web formats.}  \SWIPL comes with
  powerful interfaces to Semantic Web formats, of which we use in particular
  the XML parser and the RDF parser, which provides a call-back interface that
  allows to process in succession the triples represented in a large RDF
  document such as the \GND (see footnote~\ref{footnote-gnd} in
  Sect.~\ref{sec-introduction}).

\item \name{Parser for natural language text fragments and for formal
  languages.}  Prolog has been developed originally in the context of
  applications in linguistics and traditionally supports syntax for grammar
  rules that are translated into an advanced parsing system.  In \KBSET this
  feature is used to parse date specifications in various contexts, to parse
  person specifications by functional roles in named entity identification,
  and to implement the \LATEX parser.

\item \name{Practical workflow model.}  Workflow aspects of experimental AI
  programming seem also useful in the Digital Humanities: loading and
  re-loading documents with formal specifications as well as invocation of
  functionality and running of experiments through an interpreter.  All of
  this manageable by the researcher herself instead of further parties.

\item \name{Programming language.} Not to forget: Prolog is a programming
  language that is ``different, but not \emph{that} different''
  \cite[Introduction]{okeefe}.

\end{enumerate}

\subsection{Some Issues for Logic-Based Knowledge Processing}

\KBSET is an implemented system that has been proved workable in an
application project and allows to experimentally study further possibilities.
Some of the issues encountered in the course of implementing that were solved
in specific ways seem to deserve further investigation.
One of these issues is the interplay of knowledge that is inferred by
automated and statistic-based techniques such as named entity recognition with
manually supplied knowledge, which is addressed in \KBSET so far with the
\name{assistance documents}. Non-monotonic reasoning should in principle be a
logic-based technique that is applicable here.  Related to this issue is the
handling of \emph{ranked} query results used in \KBSET for named entity
identification. This is known in the field of databases as \name{top-k
  querying}. Is it possible to add some systematic and logic-based support for
this to Prolog and perhaps also bottom-up reasoners like deductive database
systems and model generators?

The approach to access fact bases with several millions of facts via
preprocessed caches as realized by \KBSET might be of general interest and
could be investigated and implemented more systematically.  If queries are
written in a suitable fragment of Prolog, they can be automatically optimized,
abstracting from caring about indexes (i.e., which cache is used), the order
of subgoals, and the ways in which answer components are combined.  Recent
approaches to interpolation-based query reformulation
\cite{toman:wedell:book,benedikt:book} investigate a declarative approach for
this.  The optimized version of a query is there extracted as a Craig
interpolant \cite{craig:uses,cw-pie-2020} from a proof obtained from a
first-order prover.  It seems also possible to apply this approach to
determine from a given set of queries the caches that need to be constructed
for efficient evaluation of the queries.

Digital Scholarly editing involves the interplay of natural language text with
formal code and with formalized knowledge bases.  From a general point of
view, the contribution of the computer in digital scholarly editing may be
viewed as a variant of the classical Artificial Intelligence scenario, where
an agent in an environment makes decisions on actions to perform: General
background knowledge in the AI scenario corresponds to knowledge bases like
\GND and \name{GeoNames}; the position of the agent in the environment may
correspond to a position in the text; temporal order of events to the order of
word occurrences; the environment which is only incompletely sensed or
understood by the agent corresponds to incompletely understood natural
language text; coming to decisions about actions to take corresponds to
decisions about denotations of text phrases and about annotations to associate
with text components.  This suggests that digital scholarly editing is an
interesting field for applying, improving and inventing AI techniques.

\newpage
\bibliographystyle{splncs04}
\bibliography{bib_kbset_01}

\end{document}